\documentclass[10pt,twocolumn,letterpaper]{article}

\usepackage{cvpr}
\usepackage{times}
\usepackage{epsfig}
\usepackage{graphicx}
\usepackage{amsmath}
\usepackage{amssymb}

\usepackage[linesnumbered,ruled]{algorithm2e}
\usepackage{subfigure}
\usepackage{program}
\usepackage[noend]{algpseudocode}

\usepackage[pagebackref=true,breaklinks=true,letterpaper=true,colorlinks,bookmarks=false]{hyperref}

 \cvprfinalcopy % *** Uncomment this line for the final submission

\def\cvprPaperID{2459} % *** Enter the CVPR Paper ID here

\ifcvprfinal\fi
\begin{document}

%%%%%%%%% TITLE
\title{Hyperspectral CNN Classification with Limited Training Samples}

\author{ Lloyd Windrim, Rishi Ramakrishnan, Arman Melkumyan and Richard Murphy\\
Australian Centre for Field Robotics\\
University of Sydney\\
{\tt\small \{ l.windrim,r.ramakrishnan,a.melkumyan,r.murphy \} @acfr.usyd.edu.au}
%\and
}

\maketitle

\begin{abstract}
Hyperspectral imaging sensors are becoming increasingly popular in robotics applications such as agriculture and mining, and allow per-pixel thematic classification of materials in a scene based on their unique spectral signatures. Recently, convolutional neural networks have shown remarkable performance for classification tasks, but require substantial amounts of labelled training data. This data must sufficiently cover the variability expected to be encountered in the environment. For hyperspectral data, one of the main variations encountered outdoors is due to incident illumination, which can change in spectral shape and intensity depending on the scene geometry. For example, regions occluded from the sun have a lower intensity and their incident irradiance skewed towards shorter wavelengths.

In this work, a data augmentation strategy based on relighting is used during training of a hyperspectral convolutional neural network. It allows training to occur in the outdoor environment given only a small labelled region, which does not need to sufficiently represent the geometric variability of the entire scene. This is important for applications where obtaining large amounts of training data is labourious, hazardous or difficult, such as labelling pixels within shadows. Radiometric normalisation approaches for pre-processing the hyperspectral data are analysed and it is shown that methods based on the raw pixel data are sufficient to be used as input for the classifier. This removes the need for external hardware such as calibration boards, which can restrict the application of hyperspectral sensors in robotics applications. Experiments to evaluate the classification system are carried out on two datasets captured from a field-based platform.
\end{abstract}

% STILL IN PROGRESS - RISHI 10-AUG, 23:11
\section{Introduction}
Classification algorithms which use hyperspectral data form a critical component of outdoor robotic systems as the spectral signature of an object is its most defining characteristic. While the majority of robotic platforms contain consumer grade cameras that have broadband spectral response curves, field based hyperspectral imagers are becoming increasingly common in applications such as mineral mapping of mine faces \cite{Schneider2012}, weed detection in agriculture \cite{Wendel2016}, urban imaging \cite{Ramakrishnan2015} and skin detection in search and rescue \cite{Trierscheid2008}. The chemical and structural composition of a material gives rise to unique spectral signatures, allowing per-pixel thematic classification maps of the environment to be generated through the use of supervised classification algorithms. Classifiers require labelled data to train on, but acquiring large scale ground truth labels is expensive in terms of computation and labelling effort. In such scenarios, it is desirable to label small regions and infer class labels on the remaining pixels in the image. In this work, the focus is on classification using limited amounts of training data obtained from within the image. 

Incorporating illumination variability into the training of supervised classifiers is an ongoing research question, with two main approaches being utilised. The first is by using large amounts of training data that sufficiently represent the variability within the scene. However, more data is required in order to capture such variability and this becomes increasingly difficult in complex scenes. The geometric structure of surfaces occludes regions from being illuminated evenly by terrestrial sunlight and diffuse skylight, with incident illumination and intensity varying on a per-pixel basis. The second method is to use pre-processing to convert the data to a form that is less dependant on illumination. This is typically done by converting the raw digital values to reflectance using a process by which the hyperspectral image is normalised against a material of known reflectance within the scene such as a calibration board (eg. flat-field correction). Flat-field correction is only correct for the region in which the calibration board was placed. In areas with significantly different incident illumination, the incident illumination varies based on the illumination sources and geometry of the surface. Also, placing additional hardware within the scene is impractical in hazardous environments, or for robotic platforms operating in dynamic environments. 

Convolutional Neural Networks (CNNs) have been utilised to achieve state-of-the-art performance for tasks such as classification \cite{Chatfield2014,Krizhevsky2013,Szegedy2015}. Through the use of networks pre-trained on the ImageNet database \cite{Deng2009}, CNNs have been applied to robotic applications \cite{Pinto2015,Schwarz2015}. Recently, they have also been applied to hyperspectral data captured from airborne and satellite based platforms (eg. \cite{Zhao2016}). 

In this work, the performance of CNN architectures is harnessed to generate a classification system with limited amounts of labelled training data. Illumination variability is incorporated into the classifier through the use of a data augmentation strategy that uses relighting. The advantages of the proposed approach are that it does not require multiple sensors (eg. hyperspectral camera with LiDAR) or computational atmospheric models. It allows training data to be sampled from within sunlit regions which are commonly easier to label, while classification can be performed on both sunlit and shadowed regions. The contributions of this paper are:
\begin{itemize}
\item an evaluation of several radiometric normalisation pre-processing steps,
\item the incorporation of relighting into data augmentation, 
\item an automatic, image based method for determining the incident illumination ratio, 
\item evaluation of the classification system on two field-based hyperspectral datasets captured from a field based platform.
\end{itemize}

Section~\ref{sec: prevWork} discusses related work and Section~\ref{sec: HypCNN} describes the proposed method. Experiments and results are presented in Section~\ref{sec: results}, and conclusions are drawn in Section~\ref{sec: conclusion}.

\section{Previous Work}
\label{sec: prevWork}
\subsection{Illumination Invariance in Hyperspectral Classification}
In the remote-sensing community, a well-known classifier is the Spectral Angle Mapper (SAM) which computes the inverse cosine of the normalised dot product between a spectra of interest and library spectra \cite{Yuhas1992}. SAM achieves some degree of illumination invariance as the classifier is robust to multiplications of the spectra that are constant across wavelength. Whilst this makes it invariant to different intensities, it does not cater for the effect of shadows which is a wavelength dependant multiplication of the spectra. An alternative approach is to transform the spectra into an illumination invariant form prior to classification. From computer vision, the method of \cite{Finlayson2006a} derives a transformation for RGB images to a 2D log-chromaticity space in which an axis exists where changes in illumination due to intensity and shadow are suppressed. This approach is extended to hyperspectral images in \cite{Salekdeh2011}. The problem with this approach is in the assumptions that are made in order to derive the transformation, particularly the assumption of Planckian illumination where the spectral power distribution of the incident light is modelled by Wein's approximation to Plank's law. This is because atmospheric absorption features are not accounted for by the approximation and have a large impact on the incident light spectra. Finally, multi-modal approaches \cite{Friman2011,Ramakrishnan2015} which use additional sensors such as Light Detection and Ranging (LiDAR) and Global Positioning Systems (GPS) can form geometry-based illumination models of the scene and compensate for the variations in lighting, however, the additional sensors required in these approaches are not always available.

\subsection{CNNs for Hyperspectral Classification}
In recent years, CNNs have been utilised for pixel-wise classification of hyperspectral images. However, many of the proposed CNNs convolve in the spatial domain and not the spectral domain. The CNN in \cite{Makantasis2015} convolves a small window over spatial patches extracted from a dimensionality reduced hyperspectral cube. A similar approach is taken by \cite{Zhao2016} where a CNN learns spatial features from patches extracted from a dimensionality reduced hyperspectral cube, but these spatial features are combined with spectral features learnt using a method based on Local Discriminant Embedding (LDE). These features are used together to do classification with Logistic Regression and Support Vector Machines (SVMs). In \cite{Lee2016} spatial patches are extracted from the hyperspectral cube, but this time the cubes dimensionality has not been reduced such that the patches cover the whole spectrum. This network convolves spatial filters of different sizes over the patch, again with no convolutions occurring over the spectral channel. There are some simple spatial augmentations done to avoid overfitting, but nothing to cover the large variability that can occur in the shape of the spectra.
A very simple CNN was proposed by \cite{Hu2015} which learnt features by convolving over the spectral channel. This approach was shown to perform favourably against other types of neural networks as well as SVMs. The architecture chosen consisted of only one convolutional layer and one fully-connected layer. In all of these works the datasets used for evaluation were almost exclusively captured from satellite or airborne platforms. Also, there is no analysis into the use of radiometric normalisation methods other than flat-field correction (requiring a calibration panel in the scene). Finally, all of the scenarios assume that the training examples are sufficient for capturing the variability in the data, which is not always possible in a robotics or autonomous application. 
Hence, it is unknown how robust these networks are when the training data does not capture all of the variability in the data.

\section{A Robust Hyperspectral CNN}
\label{sec: HypCNN}
In this section, a summary of how to train the network is given, followed by a description of the different pre-processing methods commonly used for hyperspectral data and the data augmentation strategy that makes the classifier robust to illumination variability. Critical to this, is the calculation of the ratio between the two primary outdoor illumination sources (terrestrial sunlight and diffuse skylight), for which a novel, image based method is proposed.
\subsection{CNN Training}
The proposed hyperspectral CNN training strategy used to increase the robustness of the classifier to illumination variability consists of multiple convolutional and non-linear activation layers, followed by fully connected layers and the output softmax classifier. Each pixel spectra is a data point and the network is trained by convolving filters over the entire spectrum. There are no pooling layers so that the location of the spectral features along the spectrum is preserved. To achieve robustness to illumination variability, each batch of training data is augmented with relighting prior to being pre-processed with radiometric normalisation and passed into the CNN.

\subsection{Radiometric Normalisation}
\label{sec: method_radnorm}
Prior to inputting the spectra into the CNN, the data can be radiometrically normalised to reduce the effects of the atmosphere and enhance spectral features. Several different methods for radiometric normalisation of hyperspectral data are included in the evaluation: 
\begin{itemize}
\item flat-field correction \cite{Rast1991} - a calibration panel of known reflectance is placed in the scene and all spectra are divided by the mean spectra across the panel. This is the most common form of radiometric normalisation.
\item residual image \cite{Marsh1983} - each spectra is scaled by a constant such that a selected channels intensity is the same as that channel's maximum across the entire scene. Then, the average intensity in each band over the entire scene is subtracted from the intensity in each channel.
\item Internal Average Relative reflectance (IARR) \cite{Kruse} - each spectra is divided by the average spectra over the entire scene.
\item continuum removal \cite{Clark1984} - a polynomial continuum across the peaks of a spectra is generated, and each spectra is divided by its continuum.
\item zero-wavelength - a constant is added to each spectra such that a chosen wavelength becomes zero.
\item raw spectra - no radiometric calibration was used.
\end{itemize} 
It is often common practice to convert the raw digital numbers from the sensor into radiance units. Having the correct units can be useful for the extraction of some manual features, but is considered a redundant process in this work and so is ignored.

\subsection{Spectral Relighting Augmentation}
\label{sec: method_dataaug}
Training supervised classification algorithms using small labelled regions fails to account for illumination variability induced by the complex geometry of a scene. This paper proposes the use of relighting as a data augmentation strategy, in order to encompass the illumination variations typically found in the outdoor environment. This allows training data to be obtained from regions in the image that are either easy to access or easy to label, and inference can then be performed on the remaining data. The following relighting derivations focus on obtaining labels from sunlit regions, and classifying on shadowed data, though the approach is easily transferable to the reverse scenario (labelling shadowed regions and inferring on sunlit regions).

The outdoor illumination model \cite{Ramakrishnan2015} consists of a parallel, terrestrial sunlight source $\mathbf{E}_{sun}\text{\boldmath$\tau$}$, and a hemispherical diffuse skylight source $\mathbf{E}_{sky}$. Assuming all materials in the scene diffusely reflect light and that indirect illumination is negligible, the radiance $L$ of a region $i$ as captured by a radiometrically calibrated camera can be approximated as:
\begin{equation}
L_i(\lambda)=\dfrac{\rho_i}{\pi}\left[ V_{i}E_{sun}(\lambda)\tau(\lambda)\cos \theta_{i} + \Gamma_{i}E_{sky}(\lambda) \right],
\label{eq: model}
\end{equation}
where $\rho_{i}$ is the albedo of the material, $V_{i}$ is a binary variable indicating whether there is line-of-sight visibility between the region and the sun position, $\theta_i$ is the angle between the surface normal $N_{i}$ and the vector towards the sun, and $\Gamma_i$ is the sky (or view) factor ranging from 0 to 1 indicating the portion of the sky dome that is visible.

Relighting is the process of simulating the spectral appearance of a region under different illumination and geometrical conditions that are not encompassed by the training set. For a single sunlit datapoint, this is achieved by multiplying the training spectra (prior to radiometric normalisation) by a wavelength dependent scaling factor:
\begin{equation}
L_{j}(\lambda) = L_{i}(\lambda) \dfrac{ V_{j}\dfrac{ E_{sun}(\lambda)\tau(\lambda) }{ E_{sky}(\lambda) } \cos \theta_{j} + \Gamma_{j} }{ \dfrac{ E_{sun}(\lambda)\tau(\lambda) }{ E_{sky}(\lambda) } \cos \theta_{i} + \Gamma_{i} },
\label{eq: scene_relighting}
\end{equation}
where $\theta_{i}$ and $\Gamma_{i}$ are the geometric parameters describing the sun angle and sky factor of the original training datapoint, while $V_j$, $\theta_{j}$ and $\Gamma_{j}$ are the parameters of the augmented datapoint. When $V_{j}$ is $0$, relighting has the effect of simulating the appearance of the original datapoint within a shadowed region, while setting it to 1 simulates the same datapoint with a different orientation. Relighting alters both the brightness and spectral curve shape of the datapoint.

In order to relight the data during training, several illumination and geometric parameters are required. The first is the terrestrial sunlight-diffuse skylight ratio $\frac{E_{sun}(\lambda)\tau(\lambda)}{E_{sky}(\lambda)}$, which describes the spectral and intensity relationship between the primary illumination sources in the outdoor environment. The geometric parameters such as the sun angle and sky factors are typically known when utilising multi-modal systems, where geo-registered point cloud data can be used explicitly to estimate these values \cite{Ramakrishnan2015}. However, for image based methods these parameters remain unknown, therefore a sampling procedure is used during training. During each batch of gradient descent optimisation of the network, the geometric parameters $\theta_A$, $V_j$, $\theta_j$, $\Gamma_j$, $\theta_i$ and $\Gamma_i$ are sampled as shown in Algorithm~\ref{alg}.

%A batch of spectra to be input into the CNN can be augmented by relighting with equations \ref{eq: scene_relighting} and \ref{eq: shadow_relighting} and randomly sampling the $\theta_{i}$ and $\Gamma_{i}$ parameters.

\begin{algorithm}
\SetKwInOut{Input}{Input}
\SetKwInOut{Output}{Output}
 \Input{batch of spectra \textbf{data}, number of irradiance ratio estimates $M$}
 \Output{augmented batch of spectra \textbf{dataAug} } 
 \textbf{dataAug} $\leftarrow$ \textbf{data} \\
 \For{k = 1 to M}{
 sample $\theta_A\sim\mathcal{U}[0,\frac{\pi}{2})$\\
 \textbf{irradianceRatio}$\leftarrow$ eq.\eqref{eq: ratio_estimate}($\theta_A$)\\
 \For{l = 1 to size of batch}{
	$V_j\sim B(1,\frac{1}{2})$,
	$\theta_i\sim\mathcal{U}[0,\frac{\pi}{2})$,
	$\theta_j\sim\mathcal{U}[0,\frac{\pi}{2}]$,
	$\Gamma_i\sim\mathcal{U}[0,1]$,
	$\Gamma_j\sim\mathcal{U}[0,1]$\\
	\textbf{relitSpectra}$\leftarrow$ eq.\eqref{eq: scene_relighting}(\textbf{data}; $V_j$,$\theta_i$,$\theta_j$,	$\Gamma_i$,$\Gamma_j$,\textbf{irradianceRatio})\\
	\textbf{dataAug} $\leftarrow$ \textbf{dataAug} $\cup$ \textbf{relitSpectra}
	}	
 }
 \caption{Augmenting a batch of spectra for training the CNN. $\mathcal{U}$ and $B$ represent uniform and Bernoulli distributions respectively.}
 \label{alg}
\end{algorithm}

%\begin{algorithm}
% \KwData{batch of spectra}
% \KwResult{augmented batch of spectra}
% M = number of irradiance ratio estimates\;
% sample \textit{M} parameters $\theta_A:U[0,\frac{\pi}{2})$ \;
% 
% \For{k=1:M}{
% $irrRatio:= eq.\eqref{eq: ratio_estimate}(\theta_A)$ \;
% sample \textit{batchsize} number of parameters\
%	$V_j:B(batchsize,\frac{1}{2})$,
%	$\theta_i:U[0,\frac{\pi}{2})$,
%	$\theta_j:U[0,\frac{\pi}{2}]$,\
%	$\Gamma_i:U[0,1]$,
%	$\Gamma_j:U[0,1]$\;
%	
%	$relitSpec_{k} := eq.\eqref{eq: scene_relighting}(Data; V_j$,$\theta_i$,$\theta_j$,$\Gamma_i$,$\Gamma_j$,$irrRatio$)\;
%	Data := [Data, $relitSpec_{k}$ ];
% }
% \caption{Augmenting a batch of spectra for training the CNN. $U$ and $B$ represent uniform and Bernoulli distributions respectively.}
% \label{alg}
%\end{algorithm}

\subsection{Image Based Estimation of the Terrestrial Sunlight-Diffuse Skylight Ratio}
\label{sec: method_ratio}

\begin{figure}[!t]
\centering
\includegraphics[width=0.4\textwidth, clip=true,trim= 0 0 0 0]{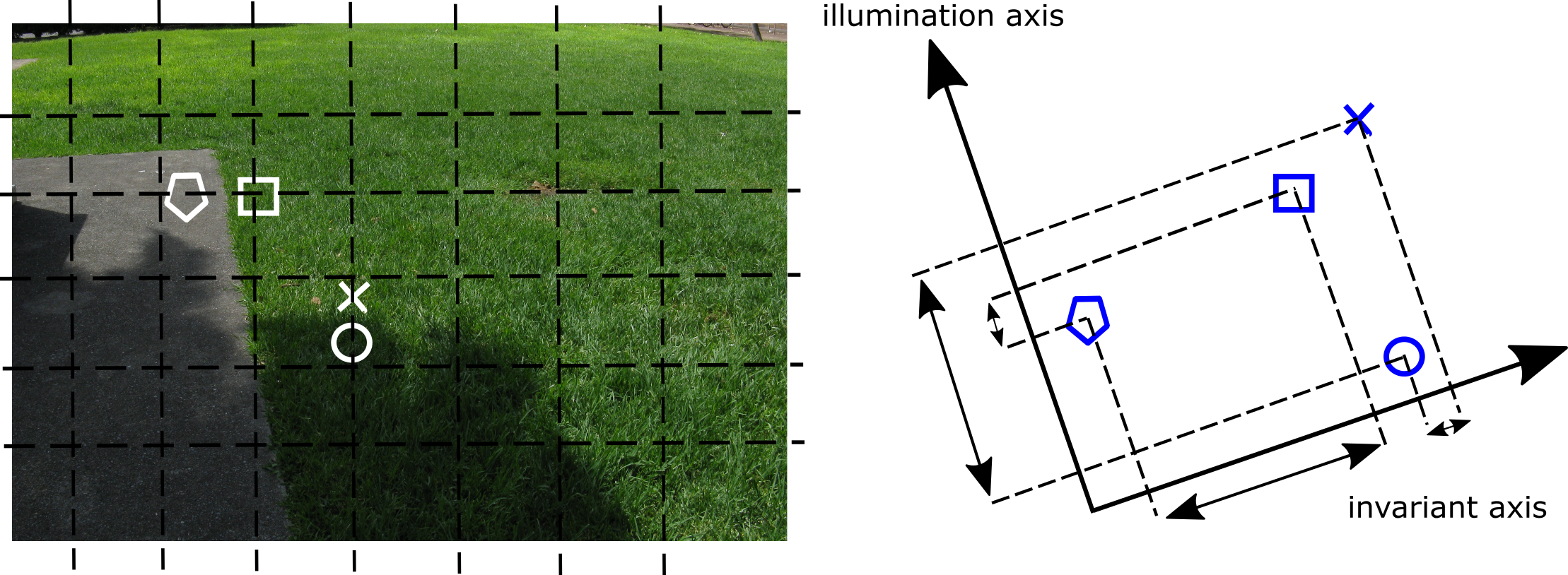}
\caption{Depiction of candidate pairs along horizontal and vertical transects projected onto illumination and invariant axes for an example image. Pairs on a class boundary have a large separation on the invariant axis, whilst pairs on a shadow boundary have a small separation on invariant axis and large separation on illumination axis.}
\label{fig:ratio}
\end{figure}

The terrestrial sunlight-diffuse skylight ratio is integral to the relighting process. Manual methods involving the user selecting two adjacent points obtained from the same material \cite{Ramakrishnan2015}, or the use of computational atmospheric models may be used, but these require the knowledge of parameters such as turbidity, gas concentration, water vapour and humidity \cite{Berk1987}. In this work, a novel image based method for estimating the terrestrial sunlight-diffuse skylight ratio from the scene is proposed.

If two pixels $(A,A')$ of the same material are selected from a sunlit and shadowed region respectively, both of which have the same orientation, the terrestrial sunlight-diffuse skylight ratio can be evaluated as:
\begin{align}
\dfrac{ E_{sun}(\lambda)\tau(\lambda) }{ E_{sky}(\lambda) } &= \dfrac{1}{\cos \theta_{A}}\left[\dfrac{L_{A}(\lambda)}{L_{A'}(\lambda)} - 1\right], \nonumber\\
& \propto \dfrac{L_{A}(\lambda)}{L_{A'}(\lambda)} - 1.
\label{eq: ratio_estimate}
\end{align}
Since the scene geometry is considered to be unknown due to the use of only image data, selection of pairs of points in sunlit and shadowed regions from the same material can be used to obtain candidate terrestrial sunlight-diffuse skylight ratios that will be a scalar multiple of the underlying ratio.

To select pairs of points in and out of the shadowed regions, three bands are generated to form a pseudo RGB image. Hypothetically these can be any bands, however they should be chosen to maximise the discriminability of the classes. If in the visible domain, these bands can be 450nm, 550nm and 600nm (peak wavelengths of an RGB camera), and if in the Short-Wave Infrared (SWIR) domain the bands can be 1060nm, 1250nm and 1630nm (the middle of sections of the spectrum outside the destructible water bands). Next, the three channel image is converted to 2D log-chromaticity space where the illumination invariant direction is found through entropy minimization \cite{Finlayson2004,Corke2013a}. The $1D$ projection of the image onto this axis should be invariant to changes in the illumination. The orthogonal axis is found (deemed the illumination axis) which captures large changes in illumination resulting from either shadow or spectrally discrete class boundaries. 

Candidate pairs of points taken from horizontal and vertical transects of the pseudo RGB image are projected onto the invariant and illumination axis:
\begin{equation}
I_{inv_{i}} = e^{X_{i}w}, 
I_{ill_{i}} = e^{X_{i}w^{\bot}},
\end{equation}
where $X_{i}$ is the log-chromaticity of point $i$, the vector $w$ is the direction of the invariant axis and $I_i$ is the exponential of the points location on either the invariant or illumination axis. If there is a large difference between a pair along the illumination axis (corresponding to either a shadow or material class boundary) but a small difference between the pair along the invariant axis (ruling out the class boundary), then the pair is considered to be valid (Figure~\ref{fig:ratio}) and the ratio between the spectra is calculated. For a candidate pair of points, the validity of them constituting a sun-shadow pair can be determined using:
\begin{equation}
\dfrac{\vert I_{inv_{1}} - I_{inv_{2}} \vert}{I_{inv_{2}}}<\mu , 
\end{equation}
\begin{equation}
\dfrac{\vert I_{ill_{1}} - I_{ill_{2}}\vert}{\min( I_{ill_{1}}, I_{ill_{2}} ) }\gt		\xi,
\end{equation}
where reasonable values for $\mu$ and $\xi$ are 0.3 and 1.2 respectively. The average ratio is subsequently taken over all valid candidate pairs before smoothing with an Savitzky-Golay filter \cite{Madden1978}. The result is used to estimate $\frac{L_{A}(\lambda)}{L_{A'}(\lambda)}$ in \eqref{eq: ratio_estimate}.

\section{Experimental Results}
\label{sec: results}
\subsection{Datasets}

% ALV picture
%\begin{figure}[!t]
%\centering
%\includegraphics[width=0.5\textwidth, clip=true,trim= 0 0 0 0]{./figures/ALV/WP_20131127_012.jpg}
%\caption{Sensor vehicle with hyperspectral camera mounted on the roof.}
%\label{fig_ALV}
%\end{figure} 

%% Great Hall
%\begin{figure*}[!t]
%\centering
%\includegraphics[width=0.8\textwidth, clip=true,trim= 80 70 80 0]{./figures/GH2.png}
%\caption{False colour image of Great Hall building with example of limited sunlit regions where labels are extracted.}
%\label{fig:GreatHall}
%\end{figure*} 

%% Great Hall classified images
%\begin{figure*}[!t]
%\centering
%\includegraphics[width=0.48\textwidth, clip=true,trim= 135 388 90 380]{./figures/GHnoaug2.pdf}
%\includegraphics[width=0.48\textwidth, clip=true,trim= 135 388 90 380]{./figures/GHaug2.pdf}
%%\includegraphics[width=0.48\textwidth, clip=true,trim= 125 388 90 360]{./figures/GHaugProb.pdf}
%%\includegraphics[width=0.48\textwidth, clip=true,trim= 125 388 90 360]{./figures/GHnoaugProb.pdf}
%%\includegraphics[width=0.48\textwidth, clip=true,trim= 135 388 90 360]{./figures/GHlabels.pdf}
%\caption{CNN classification result using \textit{left: }no augmentation and \textit{right: }augmentation. 
%%Softmax probability score for that prediction using \textit{middle left: } augmentation and \textit{middle right: } no augmentation. 
%1000 training examples per class were used, drawn from the limited regions (eg. red squares).}
%\label{fig:classGH}
%\end{figure*} 

Two field-based datasets captured from SPECIM hyperspectral sensors mounted on top of a vehicle are used to evaluate the radiometric normalisation pre-processing and proposed relighting augmentation. The first dataset is a SWIR scan of the Great Hall at the  University of Sydney (Figure~\ref{fig:classGH}) \cite{Ramakrishnan2015}.  The image consists of $293\times1306$ pixels with 152 spectral channels ranging from $1009-2482$nm. The scene consists of an urban environment, with material classes including roof, building, grass, path, tree and sky. Large shadows exist, predominately over the building and roof regions in the image due to occlusion of terrestrial sunlight. The second dataset is a Visible and Near-Infrared (VNIR) scan of a mine face \cite{Murphy2012}, exemplifying a natural, unstructured environment. It contains two types of mineral ore; Martite and Shale (Figure~\ref{fig:classMineface}). A geologist has identified a rough boundary separating the two geozones. Two scans are captured at different times of the day, one at 11:30 and one at 13:30. Each image is $289\times1443$ pixels, with 220 spectral channels corresponding to wavelengths in the range of 401-970nm. The image captured during the afternoon shows the emergence of shadows due to occlusion of terrestrial sunlight.

\subsection{Experimental Method and Metrics}

Three experiments are conducted to evaluate different components of the proposed classification system, namely, the radiometric normalisation pre-processing, CNN architecture, and relighting augmentation.

Both datasets are used in all three experiments (only the 13:30 image of the mine face is used for Sections~\ref{sec: results_radio} and \ref{sec: results_arch}). To assess the performance of the CNN for each of the experiments, the Area Under Curve (AUC) and F1 score is used. The CNN assigns a probability score distributed over all of the classes. Thus, a threshold probability score should be selected to determine whether a class is assigned to a pixel or not (for very low probabilities it is better not to assign a class). The AUC score looks at the area underneath the precision-recall curve for all threshold values whilst the F1 score is calculated by selecting a threshold that maximises the F1 score on a labelled validation dataset. In all experiments, test set sizes of roughly 225,000 pixel spectra for the Great Hall dataset and 90,000 pixel spectra for the mine face dataset are used to evaluate the classifiers. The validation dataset consists of 50 examples per class. The mean and standard deviation for five repetitions is recorded.

The overall CNN architecture is the same for both datasets, with the first convolutional layer learning 30 filters and always having a filter size of $1\times 30\times 1$ (each spectra was reshaped to be $1\times D\times 1$ where $D$ is the number of spectral channels). In subsequent convolutional layers, ten filters are learnt each with a size of $1\times 10$. Fully connected layers always consist of 20 units. The networks are optimised using 200 epochs of Stochastic Gradient Descent with a learning rate of $10^{-5}$, momentum of 0.9 and batch size of 50. For each repetition, networks are randomly initialised.

\subsection{Radiometric Normalisation}
\label{sec: results_radio}
The various types of radiometric normalisation methods used as a pre-processing step prior to input of the hyperspectral data into the CNN are evaluated. Training data is sampled from the entire scene, covering sunlit and shaded regions, so that sufficient coverage of the scene geometry and incident illumination can be made. The number of training examples per class is 500, and a two convolutional layer and two fully connected layer architecture is used. No data augmentation is used in this experiment.

\begin{table*}[htbp]
\footnotesize
  \centering
  \caption{Radiometric Normalisation Results: Classification scores for Great Hall}
    \begin{tabular}{l|rrrrrrr|r} \hline
          & \multicolumn{7}{c}{\textbf{F1 score}} & \multicolumn{1}{|l}{\textbf{AUC}} \\ 
    \textbf{Normalisation Method} & \multicolumn{1}{|l}{\textbf{Roof}} & \multicolumn{1}{l}{\textbf{Building}} & \multicolumn{1}{l}{\textbf{Grass}} & \multicolumn{1}{l}{\textbf{Path}} & \multicolumn{1}{l}{\textbf{Tree}} & \multicolumn{1}{l}{\textbf{Sky}} & \multicolumn{1}{l}{\textbf{Mean}} & \multicolumn{1}{|l}{\textbf{Mean}} \\ \hline
    Residual Image & 91.43$\pm$0.83 & 98.06$\pm$0.46 & 99.49$\pm$0.14 & \textbf{99.07$\pm$0.24} & 97.93$\pm$0.54 & 99.21$\pm$0.24 & 97.53$\pm$0.16 & 99.91$\pm$0.01 \\
    IARR  & 86.61$\pm$0.45 & 96.27$\pm$0.42 & 99.62$\pm$0.03 & 94.69$\pm$0.54 & 98.36$\pm$0.73 & 99.31$\pm$0.10 & 95.81$\pm$0.25 & 99.67$\pm$0.03 \\
    Continuum Removal & 66.85$\pm$2.39 & 95.19$\pm$0.57 & 98.05$\pm$0.24 & 85.61$\pm$3.34 & 98.68$\pm$0.15 & 98.04$\pm$0.58 & 90.40$\pm$0.86 & 99.06$\pm$0.07 \\
    Zero-wavelength & 92.80$\pm$0.72 & 98.14$\pm$0.52 & 99.73$\pm$0.04 & \textbf{99.07$\pm$0.27} & 99.44$\pm$0.12 & \textbf{99.61$\pm$0.07} & 98.13$\pm$0.18 & \textbf{99.97$\pm$0.01} \\
    Flat-field & 75.01$\pm$0.52 & 94.57$\pm$0.32 & 99.59$\pm$0.04 & 92.99$\pm$0.33 & 96.93$\pm$0.92 & 98.59$\pm$0.33 & 92.95$\pm$0.32 & 99.49$\pm$0.03 \\
    Raw Spectra & \textbf{95.02$\pm$0.30} & \textbf{98.90$\pm$0.40} & \textbf{99.74$\pm$0.17} & 99.06$\pm$0.34 & \textbf{99.48$\pm$0.08} & 99.53$\pm$0.06 & \textbf{98.62$\pm$0.15} & \textbf{99.97$\pm$0.02} \\ 
    \end{tabular}%
  \label{tab:calibGreathall}%
\end{table*}%

\begin{table*}[htbp]
\footnotesize
  \centering
  \caption{Radiometric Normalisation Results: Classification scores for mine face}
    \begin{tabular}{l|rrr|r} \hline
          & \multicolumn{3}{c}{\textbf{F1 score}} & \multicolumn{1}{|l}{\textbf{AUC}} \\ 
    \textbf{Normalisation Method} & \multicolumn{1}{|l}{\textbf{Martite }} & \multicolumn{1}{l}{\textbf{Shale}} & \multicolumn{1}{l}{\textbf{Mean }} & \multicolumn{1}{|l}{\textbf{Mean}} \\ \hline
    Residual Image & 99.83$\pm$0.03 & \textbf{99.87$\pm$0.02} & \textbf{99.85$\pm$0.02} & \textbf{100.00$\pm$0.00} \\
    IARR  & 98.29$\pm$1.40 & 98.34$\pm$1.76 & 98.32$\pm$1.58 & 99.95$\pm$0.11 \\
    Continuum Removal & 99.37$\pm$0.03 & 99.50$\pm$0.02 & 99.44$\pm$0.03 & 99.97$\pm$0.00 \\
    Zero-wavelength & \textbf{99.84$\pm$0.07} & \textbf{99.87$\pm$0.06} & \textbf{99.85$\pm$0.06} & \textbf{100.00$\pm$0.00} \\
    Flat-field & 80.75$\pm$7.78 & 83.20$\pm$5.68 & 81.98$\pm$6.73 & 90.80$\pm$4.80 \\
    Raw Spectra & 99.81$\pm$0.05 & 99.85$\pm$0.04 & 99.83$\pm$0.04 & \textbf{100.00$\pm$0.00} \\ 
    \end{tabular}%
  \label{tab:calibMining}%
\end{table*}%

\begin{table*}[htbp]
\footnotesize
  \centering
  \caption{CNN Architecture Results: Classification scores for Great Hall}
    \begin{tabular}{rr|rrrrrrr|r} \hline 
          &       & \multicolumn{7}{c}{\textbf{F1 score}} & \multicolumn{1}{|l}{\textbf{AUC}} \\ 
    \multicolumn{1}{l}{\textbf{\# conv layers}} & \multicolumn{1}{l|}{\textbf{\# fc layers}} & \multicolumn{1}{l}{\textbf{Roof}} & \multicolumn{1}{l}{\textbf{Building}} & \multicolumn{1}{l}{\textbf{Grass}} & \multicolumn{1}{l}{\textbf{Path}} & \multicolumn{1}{l}{\textbf{Tree}} & \multicolumn{1}{l}{\textbf{Sky}} & \multicolumn{1}{l|}{\textbf{Mean}} & \multicolumn{1}{l}{\textbf{Mean}} \\ \hline
    1     & 2     & 91.10$\pm$2.52 & 97.64$\pm$1.13 & \textbf{99.72$\pm$0.11} & 96.85$\pm$2.64 & 99.28$\pm$0.28 & 99.35$\pm$0.36 & 97.32$\pm$1.10 & 99.83$\pm$0.16 \\
    2     & 1     & 88.33$\pm$1.36 & 97.42$\pm$0.57 & 99.65$\pm$0.12 & 97.89$\pm$0.65 & 99.18$\pm$0.21 & 99.48$\pm$0.10 & 96.99$\pm$0.33 & 99.89$\pm$0.03 \\
    2     & 2     & 93.25$\pm$1.50 & 98.03$\pm$0.84 & 99.70$\pm$0.10 & 98.60$\pm$1.13 & 99.41$\pm$0.08 & 99.49$\pm$0.09 & 98.0$\pm$0.44 & 99.95$\pm$0.02 \\
    2     & 3     & \textbf{93.48$\pm$0.70} & \textbf{98.20$\pm$0.62} & 99.69$\pm$0.09 & \textbf{98.86$\pm$0.46} & 99.41$\pm$0.07 & 99.48$\pm$0.12 & \textbf{98.19$\pm$0.22} & 99.94$\pm$0.03 \\ 
    3     & 2     & 92.62$\pm$2.55 & 97.96$\pm$0.81 & 99.70$\pm$0.04 & 98.84$\pm$0.83 & \textbf{99.47$\pm$0.05} & \textbf{99.56$\pm$0.07} & 98.02$\pm$0.66 & \textbf{99.96$\pm$0.02} \\
    \end{tabular}%
  \label{tab:archGreathall}%
\end{table*}%

The results from the Great Hall (Table~\ref{tab:calibGreathall}) and mine face (Table~\ref{tab:calibMining}) datasets  shows that the zero-wavelength radiometric normalisation and raw spectra methods obtained the best results. This suggests that less pre-processing is actually beneficial for training a CNN. Further, the flat-field approach did not perform well in comparison to several of the other methods. In the remote sensing community it has been common practice to use flat-field correction with a calibration panel placed in the scene to convert hyperspectral data to a reflectance feature space that is more discriminant than the raw spectra. This work has suggested that the flat-field correction approach does not always produce the best feature space for CNN classification, and that it is more beneficial to have the CNN learn the features from the data. 

The zero-wavelength approach will be used in the hyperspectral classification system. Unlike flat-field correction, no physical interaction with the scene is required which is important for robotics and autonomous applications where the dynamic nature of the sensor vehicle means that placing calibration panels in the scene is often impractical.

\subsection{CNN Architecture}
\label{sec: results_arch}
Five different CNN architectures are evaluated while keeping the pre-processing method fixed. The same training data is used as in Section~\ref{sec: results_radio}, with the purpose of the experiment being to analyse the impact of using multiple convolutional and fully connected layers.

Table~\ref{tab:archGreathall} reports the Great Hall dataset results and suggests that little benefit is obtained in using more than two convolutional layers. Similarly, there is minimal gain when moving from two to three fully connected layers. Hence, for computation and performance, two convolutional layers and two fully connected layers are considered appropriate for the classification system for both datasets.

%The results from the experiments comparing architectures (Table.~\ref{tab:archGreathall} reports the Great Hall dataset results) suggests that there is very little difference in the results using different architectures, particularly with two or more  convolutional and fully connected layers. Hence, for computation and performance, two convolutional layers and two fully connected layers are considered appropriate for both datasets.

\subsection{Spectral Relighting Augmentation} 
\label{sec: results_aug}

% Great Hall classified images
\begin{figure*}[!t]
%\hfill
\centering
\subfigure[False colour image of Great Hall building with example of limited sunlit regions where labels are extracted.]{\includegraphics[width=0.55\textwidth, clip=true,trim= 80 70 80 0]{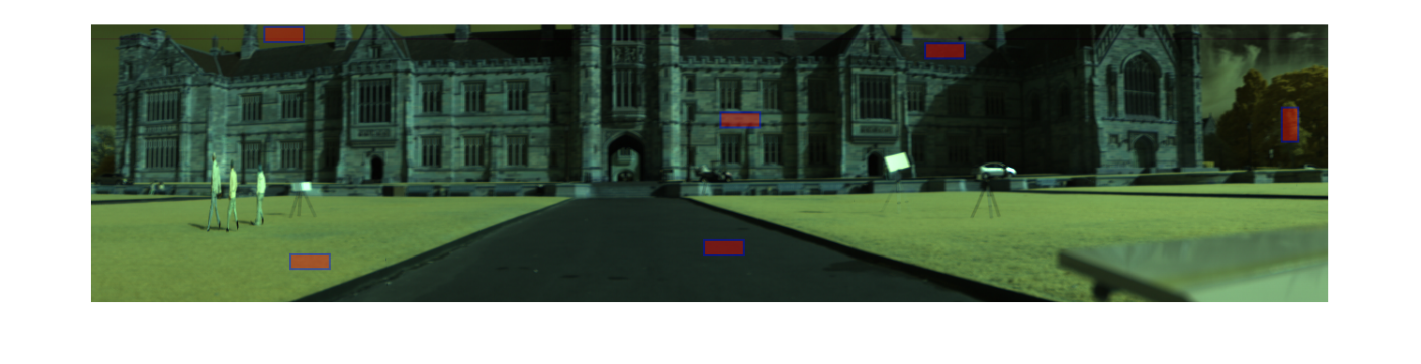}}\\
\hfill 
\subfigure[No augmentation.]{\includegraphics[width=0.49\textwidth, clip=true,trim= 135 388 90 380]{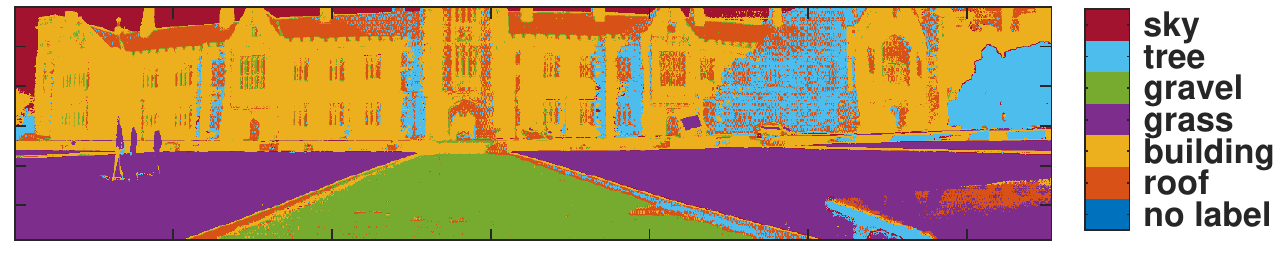}}
\hfill
\subfigure[Augmentation]{\includegraphics[width=0.49\textwidth, clip=true,trim= 135 388 90 380]{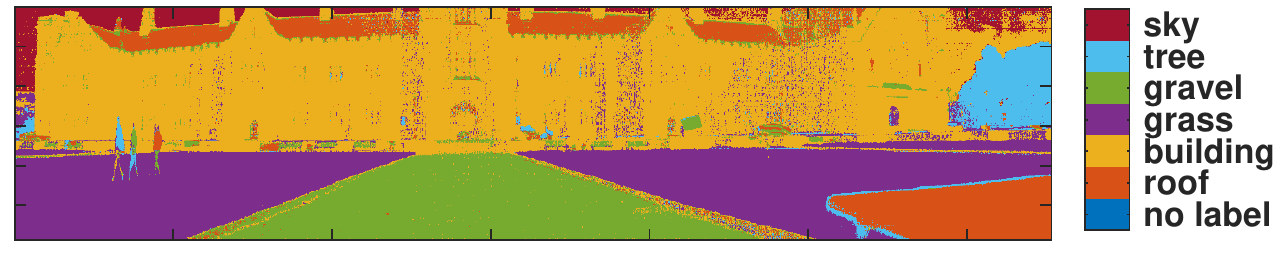}}
\hfill
\caption{Spectral Relighting Augmentation Results: CNN classification of the Great Hall using 1000 training examples per class were used, drawn from the limited regions (eg. red squares).}
\label{fig:classGH}
\end{figure*}

\begin{figure*}[!t]
\hfill
\subfigure[Mine face image captured at 11:30]{\includegraphics[width=0.49\textwidth, clip=true,trim= 0 0 0 0]{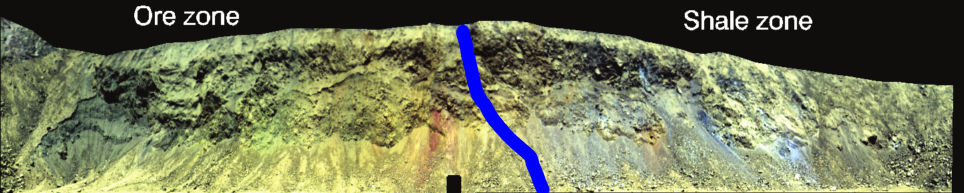}}
\hfill
\subfigure[Mine face image captured at 13:30 with example of limited sunlit regions where labels are extracted from.]{\includegraphics[width=0.49\textwidth, clip=true,trim= 0 0 0 0]{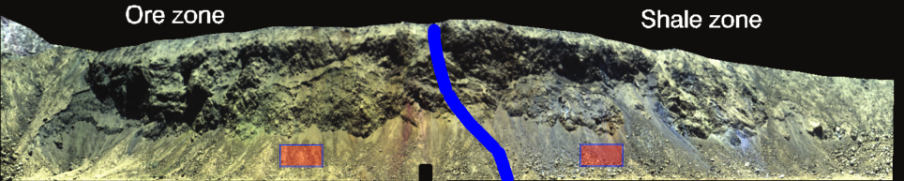}} 
\hfill 
\subfigure[No augmentation 11:30]{\includegraphics[width=0.49\textwidth, clip=true,trim= 132 395 90 380]{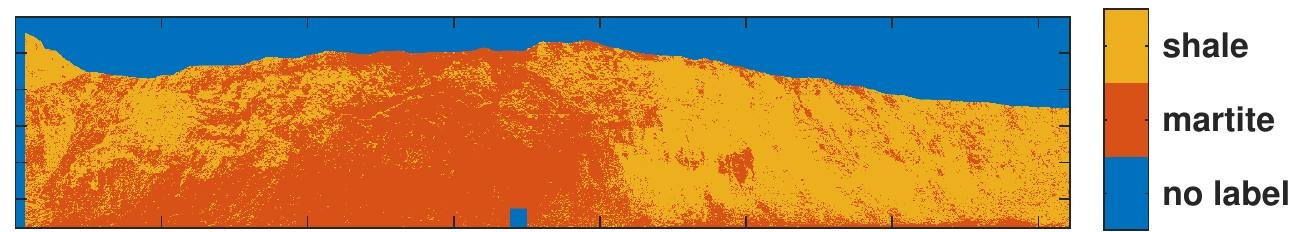}}
\hfill
\subfigure[No augmentation 13:30]{\includegraphics[width=0.49\textwidth, clip=true,trim= 132 395 90 380]{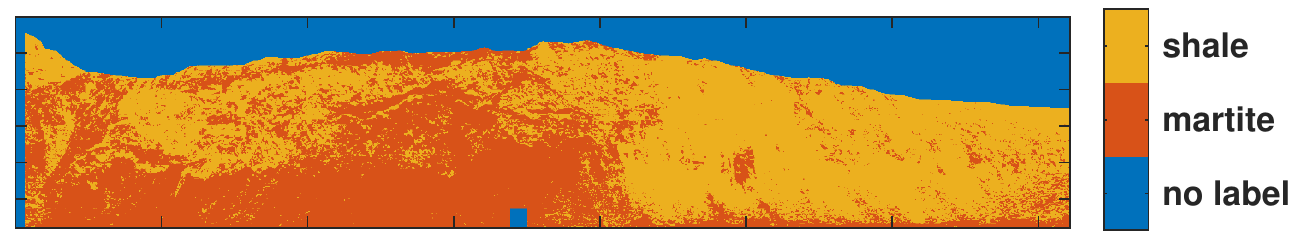}}
\hfill
\subfigure[Augmentation 11:30]{\includegraphics[width=0.49\textwidth, clip=true,trim= 132 395 90 380]{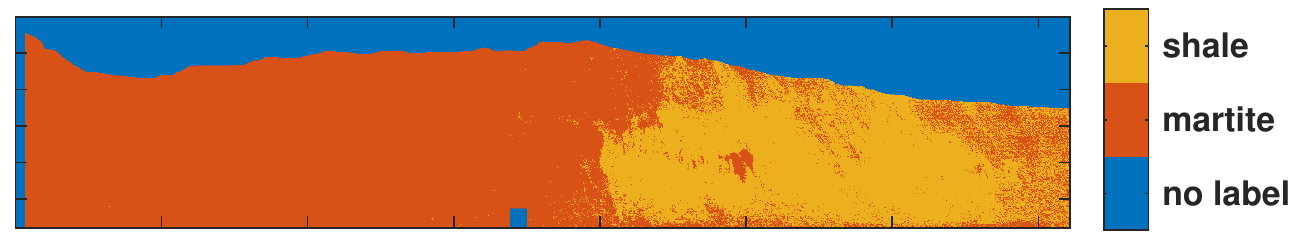}}
\hfill
\subfigure[Augmentation 13:30]{\includegraphics[width=0.49\textwidth, clip=true,trim= 132 395 90 380]{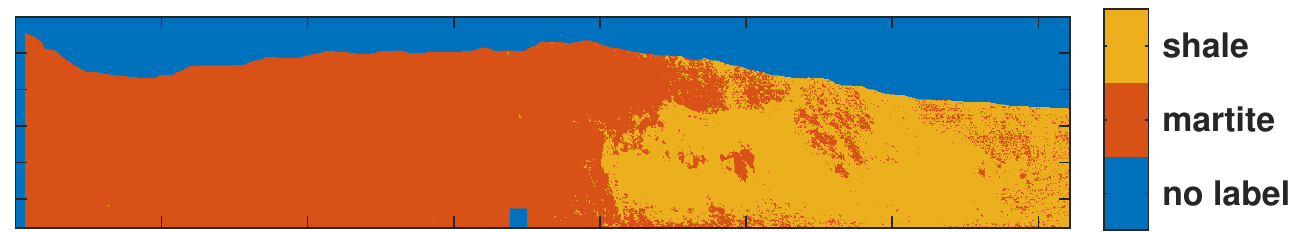}}
\hfill
\caption{Spectral Relighting Augmentation Results: CNN classification of the mine face using 100 training examples per class were used, drawn from the limited regions (eg. red squares). Note that the same network was used to classify the image at both times of the day.}
\label{fig:classMineface}
\end{figure*}

\begin{figure}
\hfill
\subfigure[Great Hall]{\includegraphics[width=0.47\textwidth, clip=true,trim= 70 260 70 282]{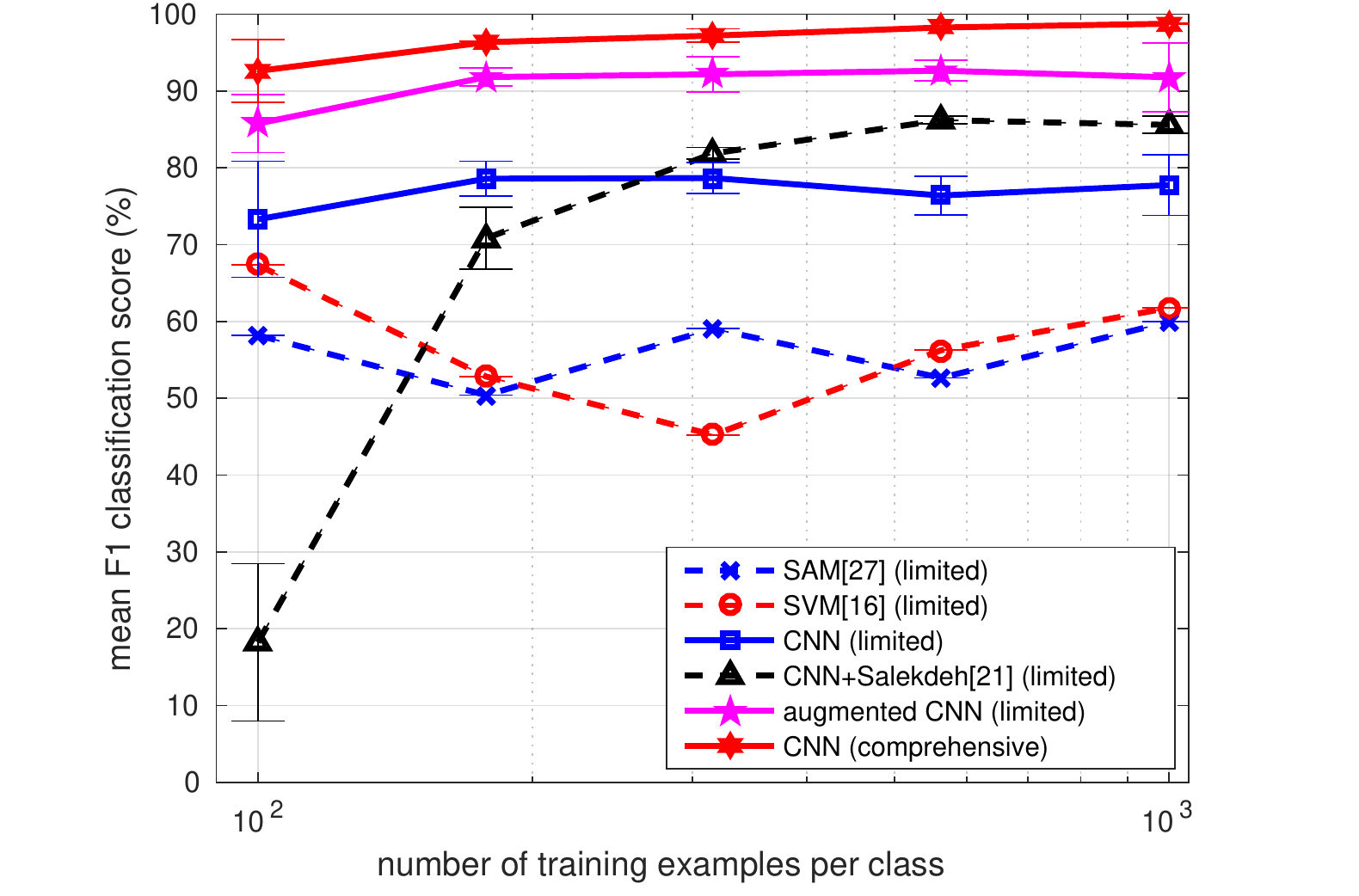}}
\hfill
\subfigure[Mine face]{\includegraphics[width=0.47\textwidth, clip=true,trim= 70 260 70 282]{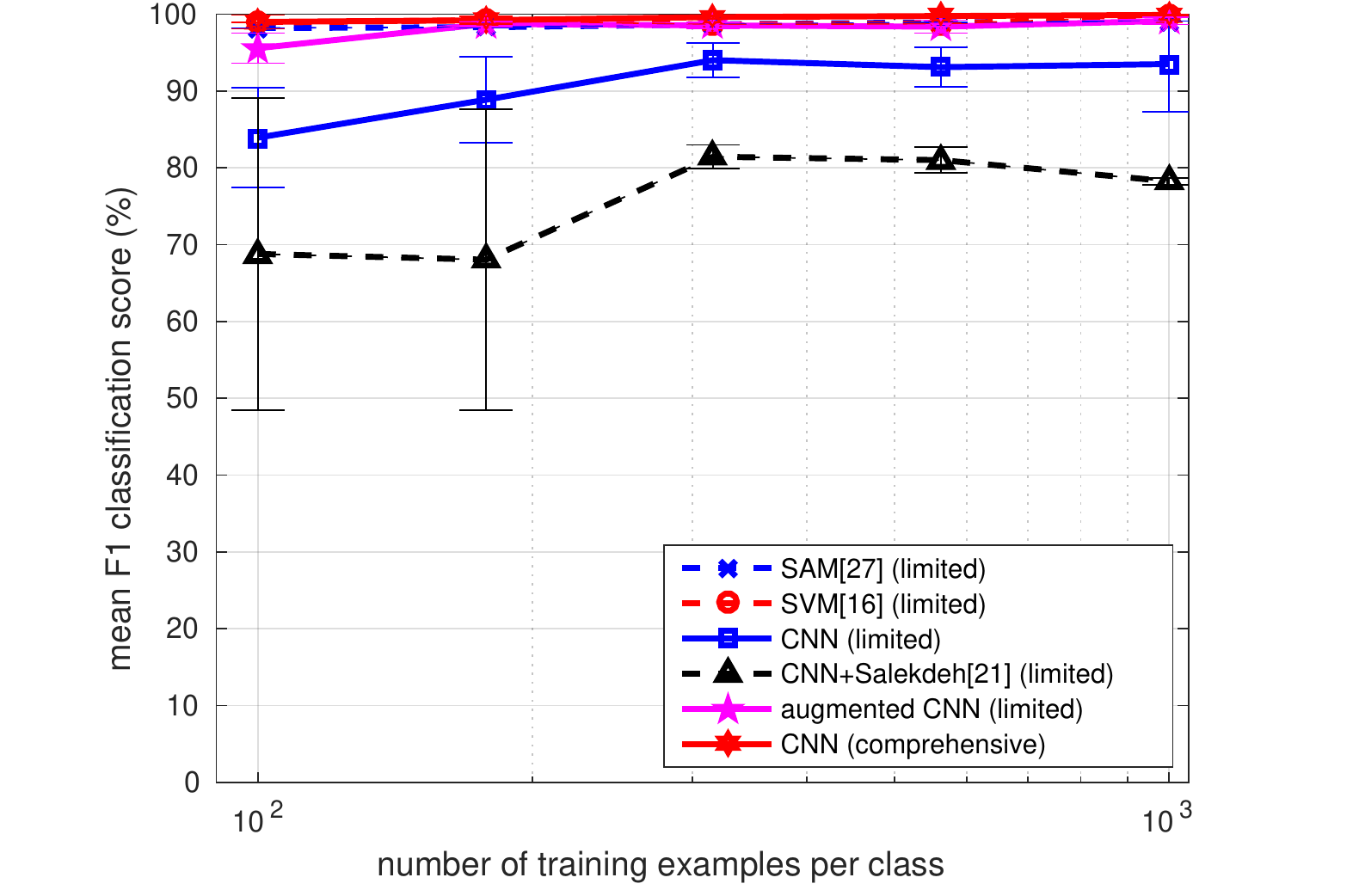}} 
\hfill \\
\caption{Spectral Relighting Augmentation Results: Mean and standard deviation of the F1 classification score for several different methods, trained on limited regions and the whole image (comprehensive), for different sized training sets.}
\label{fig:augplot}
\end{figure}

\begin{figure}
\hfill
\subfigure[Roof class]{\includegraphics[width=0.47\textwidth, clip=true,trim= 70 260 70 282]{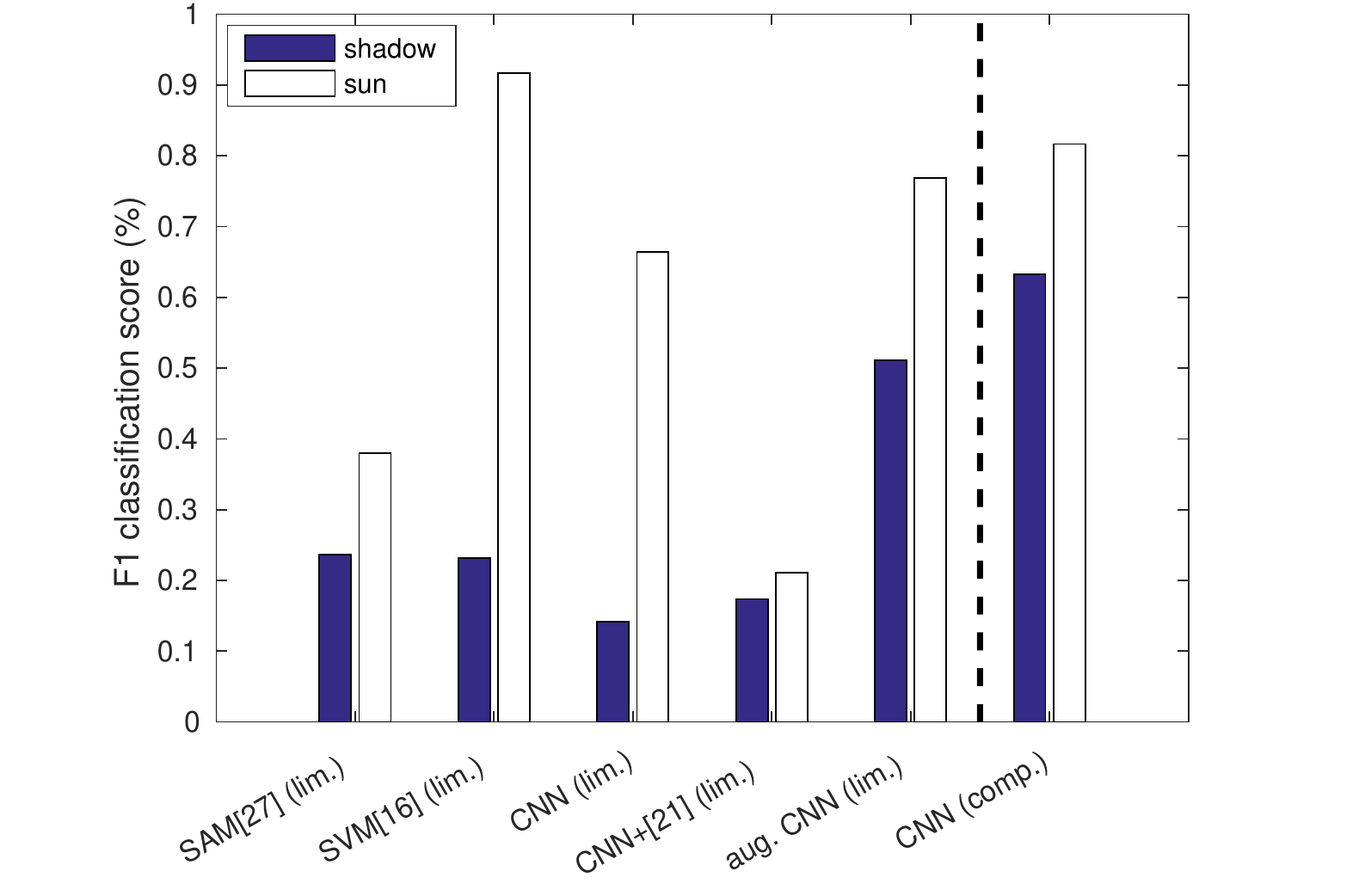}}
\hfill
\subfigure[Building class]{\includegraphics[width=0.47\textwidth, clip=true,trim= 70 260 70 282]{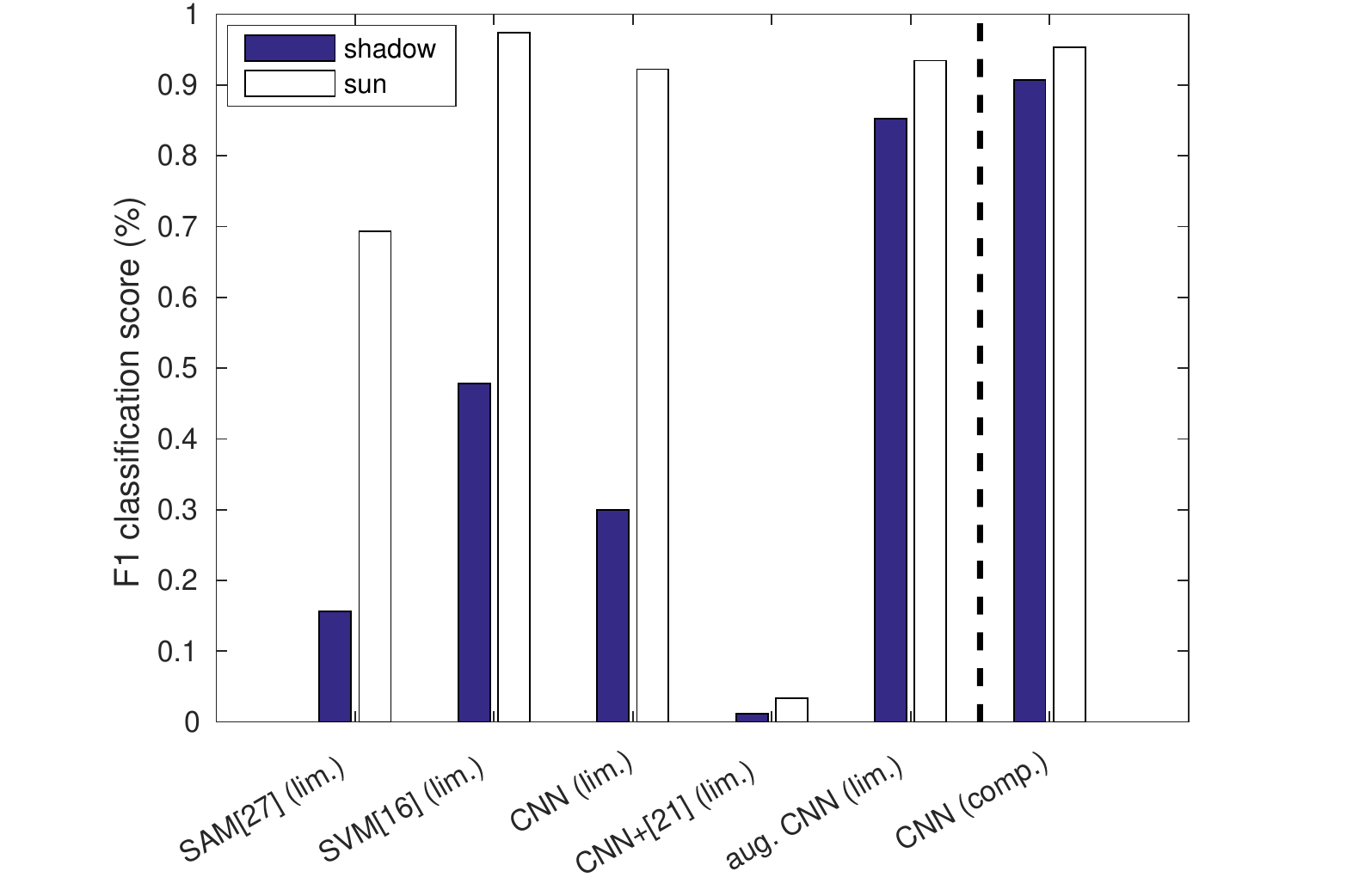}} 
\hfill \\
\caption{Spectral Relighting Augmentation Results: F1 classification score of Great Hall classes in shadow and sunlight, for several different methods, trained on limited regions and the whole image (comprehensive),  for training set size 100 samples per class.}
\label{fig:augShadowGreatHall}
\end{figure}

The proposed relighting augmentation is evaluated using a fixed pre-processing method and architecture selected from the results of Sections~\ref{sec: results_radio} and \ref{sec: results_arch}. Two sets of labelled training data are used. One set (referred to as `comprehensive') is collected from both sunlit and shaded regions of the image, and also has large spatial coverage such that it best represents the variation in scene geometry and incident illumination. The second labelled training dataset (referred to as `limited') is only collected in sunlit regions and the spatial coverage is small - limited to a patch (red squares in Figures~\ref{fig:classGH} and \ref{fig:classMineface}), such that there is a very poor representation of the scene geometry and incident illumination. Networks are trained using no augmentation on both the limited and comprehensive datasets and also using augmentation on the limited dataset. The results of the networks trained on the comprehensive training data can be seen as an upper bound for the results of training on the limited datasets. The performance of the CNN trained on the limited data, both augmented and not augmented, is compared against other classification approaches including SAM \cite{Yuhas1992}, an SVM \cite{Melgani2004}, and a CNN trained on spectra projected into an illumination invariant space using log-chromaticity \cite{Salekdeh2011}. For the mine face dataset, networks are trained on the 13:30 image only, since it contains shadowed regions. 

The number of training examples is varied logarithmically between 100 and 1000 examples per class (5 increments: 100, 178, 316, 562, 1000). Ten candidate irradiance ratios are generated for the augmentation such that each pixel spectra is relit ten times, with roughly half of those relightings being to shadow and half remaining sunlit but with different orientations. This expands the training dataset to 11 times its original size.

The results (Figure~\ref{fig:augplot}) from both the Great Hall and mine face datasets show that there was a clear advantage to augmenting the labelled data used to train the CNN. The CNN achieved better classification scores with augmentation in comparison to not using augmentation for all training set sizes. By augmenting the training data, most of the gap between training the CNN with limited and comprehensive training sets was bridged. The results of training with the augmented data was also superior to the CNN trained on data that was projected into an illumination invariant space \cite{Salekdeh2011}. The SAM and SVM classifiers performed comparably on the mine face dataset, which only had two classes, but performed significantly worse on the Great Hall dataset which had six classes, suggesting that their use may be limited to scenarios with a small number of classes. The performance of the augmented and non-augmented CNN as the amount of training samples was increased was relatively consistent. This is because the training examples were being sampled from such a small region of the image such that increasing them allowed the CNN to capture very little extra variability. Since the augmented CNN simulated the missing variability there was little dependence of the classifier's performance on the number of training examples.

Figure~\ref{fig:augShadowGreatHall} shows the improvement that augmented training data provided specifically in the shadowed and sunlit areas for two of the Great Hall classes. In both sun and shadow, most of the gap between training the CNN with limited and comprehensive data was bridged by the augmentation, suggesting that it allowed the limited amount of training data to capture the variability as if the labels were collected from all over the image - covering areas with different geometry, incident illumination and occlusions. Also, in contrast to the SVM, which performed well on the sunlit regions for these two classes but poorly on the shadowed regions, the augmented CNNs performance in the shadow approached its performance in the sun. The CNN trained on the illumination invariant projection of the data \cite{Salekdeh2011} had similar performance in sunlit and shadowed regions but was significantly worse than the augmented CNN, whose performance in the shadow surpassed all other methods trained on limited data. Despite the improvement, Figure~\ref{fig:classGH} shows that there are still some sporadic misclassifications in the shadowed regions which could be due to indirect illumination.

%For two of the Great Hall classes, Fig.~\ref{fig:augShadowGreatHall} shows the improvement that augmented training data provided specifically in the shadowed and sunlit areas. In both sun and shadow, most of the gap between training the CNN with limited and comprehensive data was bridged by doing the augmentation, suggesting that it allowed the limited amount of training data to capture the variability as if the labels were collected from all over the image - covering areas with different geometry, incident illumination and occlusions. Also, in contrast to the SVM, which performed well on the sunlit regions for these two classes but poorly on the shadowed regions, the augmented CNNs performance in the shadow approached its performance in the sun. The CNN trained on the illumination invariant projection of the data \cite{Salekdeh2011} had similar performance in sunlit and shadowed regions of the image but both were a significantly worse result than the augmented CNN, whose performance in the shadow surpassed all other methods trained on the limited data. Despite the improvement, Fig.~\ref{fig:classGH} shows that there are still some sporadic misclassifications in the shadowed regions which could be due to indirect illumination. This can be improved by incorporating indirection illumination into the model.

% augmentation Mining
\begin{figure}[!t]
\centering
\includegraphics[width=0.47\textwidth, clip=true,trim= 70 260 70 282]{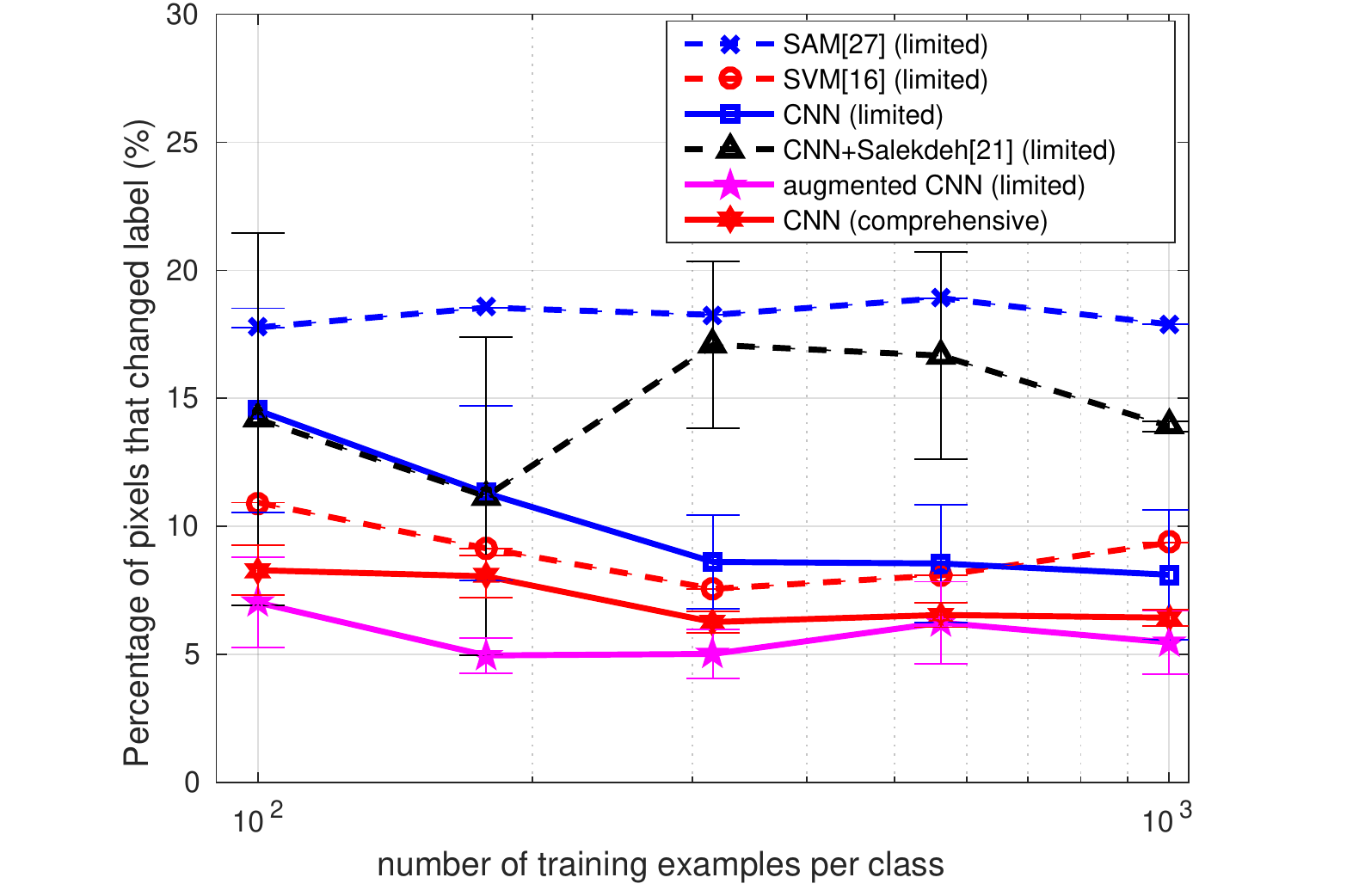}
\caption{Spectral Relighting Augmentation Results: Percentage of pixels that changed classification label from the 11:30 to 13:30 mine face images for several different methods trained on localised regions and the whole image  for different sized training sets.}
\label{fig:augPercChanceMineface}
\end{figure}

Figure~\ref{fig:classMineface} shows that the CNN trained with augmented data produced a better result in the shadow than the CNN trained without augmentation as the classification was more reflective of the ground truth geozones. The CNN trained with augmentation also exhibited greater temporal invariance over the day than all other approaches, given by the reduced percentage of pixels that changed classification label between 11:30 and 13:30 (Figure~\ref{fig:augPercChanceMineface}). Note that this CNN would not be invariant over different days as the terrestrial sunlight-diffuse skylight ratio would change.% due to variable atmospheric conditions.

\section{Conclusion}
\label{sec: conclusion}
This work has proposed a method for training a CNN to be robust to several factors of variation in hyperspectral data with only a limited amount of training data. This was done through relighting augmentation using an approximation of the irradiance ratio found from the scene.  This work also showed that it is possible to train a CNN to classify hyperspectral data without first radiometrically normalising it using a calibration panel in the scene. This has paved the way for utilisation of CNNs with hyperspectral cameras in robotics applications. Future work involves the incorporation of spatial information and developing large scale, multi-label classifiers.

\section{Acknowledgments}
This work has been supported by the Rio Tinto Centre for Mine Automation and the Australian Centre for Field Robotics, University of Sydney.
%\clearpage

{\small
\bibliographystyle{ieee}
\bibliography{ICRA2017}
}

\end{document}